\documentclass[letterpaper, 10 pt, conference]{ieeeconf}  
\IEEEoverridecommandlockouts                              
\usepackage{graphicx}
\usepackage{bm}
\usepackage{cite}
\usepackage{multirow}
\usepackage{threeparttable}
\usepackage{hyperref}

\title{\LARGE \bf
Pouring by Feel: An Analysis of Tactile and Proprioceptive Sensing for Accurate Pouring
}

\author{Pedro Piacenza, Daewon Lee and Volkan Isler
\thanks{All authors are with Samsung AI Center NY, 837 Washington Street New York, New York 10014
}%
}

\begin{document}
\maketitle
\thispagestyle{empty}
\pagestyle{empty}
\setlength{\parskip}{0pt}

\begin{abstract}

As service robots begin to be deployed to assist humans, it is important for them to be able to perform a skill as ubiquitous as pouring. Specifically, we focus on the task of pouring an exact amount of water without any environmental instrumentation, that is, using only the robot's own sensors to perform this task in a general way robustly. In our approach we use a simple PID controller which uses the measured change in weight of the held container to supervise the pour. Unlike previous methods which use specialized force-torque sensors at the robot wrist, we use our robot joint torque sensors and investigate the added benefit of tactile sensors at the fingertips. We train three estimators from data which regress the poured weight out of the source container and show that we can accurately pour within 10 ml of the target on average while being robust enough to pour at novel locations and with different grasps on the source container. Video: \url{https://youtu.be/fQw6i8FpENI}

\end{abstract}
\section{Introduction}

Pouring is a common daily task, especially in the context of food preparation, housekeeping, bartending and hospitality services. As service robots begin to be deployed to assist humans in various environments, it is important for general purpose robots to be able to accurately pour a specific amount of liquid into a container. 


Robotic pouring has been widely demonstrated using different modalities such as vision\cite{schenck2017visual,zhang2021explainable}, audio\cite{liang2019making} and haptics\cite{liang2020robust,rozo2013force} involving varying levels of setup instrumentation and structure. Inspired by human pouring, vision seems to be the more relevant sensor modality. However, the perception task of determining the current liquid level in the receiving container is far from trivial, and most work in this area requires controlled conditions. Real world scenarios where containers may be opaque, liquid color varies, background and lighting are not controlled, and the camera point of view might be less-than-ideal complicate this method even further.

Another common approach is the use of force sensors at the robot wrist or underneath the receiving container to measure the weight of the poured liquid. In some contexts, such as a kitchen robot, it may be acceptable to designate a specialized area equipped with a force sensor to assist in the pouring task. On the other hand, a mobile service robot would be expected to perform this task without such instrumentation (for example, for a human holding the glass). 

For this reason we are interested in investigating how accurately we can pour liquids without \textit{any} environmental instrumentation such as fixed cameras or force sensors underneath receiving containers. We envision a robot that can approach a table and pour drinks accurately and reliably using only its own sensors. More specifically we focus on the ability to measure how much weight has been poured out of the held container by using the robot arm joint torque sensors supplemented with tactile sensation at the fingertips. 

\begin{figure}[t]
\centering
\includegraphics[clip, trim=7cm 1.5cm 10cm 3.5cm, width=0.8\linewidth]{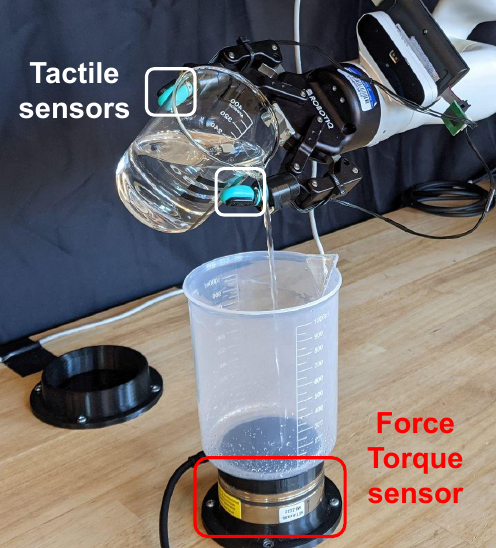}
\caption{Our experimental setup uses a Kinova Gen3 arm with joint torque sensors and a Robotiq 2F-85 gripper equipped with BioTac tactile sensors. During training trials a PID controller reads the weight at the receiving container from a force-torque sensor to regulate the pouring rate}
\label{fig:exp_setup}
\end{figure}

Unlike previous work, we intentionally decide against the use of a dedicated force-torque sensor at the wrist to avoid the extra payload and wiring in its integration into the manipulator. Instead, we focus on using the robot arm's built-in joint torque sensors, as this capability is becoming more popular and widely available in commercial robot arms. This makes the task more challenging, as the measurements provided by joint torque sensors are noisier and less accurate compared to those of a dedicated force-torque sensor. Moreover, joint torques do not provide a direct measurement of forces at the robot end-effector. To bridge this potential gap in performance, we introduce the use of tactile sensors at the fingertips, which relay information about the container weight through skin deformation caused by shear forces. 

Human studies have shown that skin deformation at the fingertips is a big contributor to how we perceive the weight of objects, especially for light ones \cite{kamikawa2018comparison, kayhan2018skin, matsui2013relative}. Our hypothesis is that we can take advantage of tactile signals proportional to skin deformation to compensate for the lack of detail in our joint torque signals, and effectively determine the change in weight at the source container.

To study the use of torque and tactile sensors in the context of pouring, we start by training three estimators: one using mainly tactile information, another using the joint torques information and the last one combining the two sensing modalities. These estimators can regress the weight poured out of the held container at any given moment. To train them, we collect pouring trials using a simple PID controller which regulates the pouring rate by adjusting the wrist velocity. The controller's feedback is provided by a force sensor placed underneath the receiving container which is only needed at training time. Assuming all weight lost at the held container arrives later at the receiving container with no spillage, we can adjust the ground truth signal by shifting it in time to represent the weight as it leaves the held container and relate it to our sensing modalities of interest. 

We evaluate our estimators performance in two ways: first, we compute their mean squared error against ground truth on our test set, and second, we measure their pouring accuracy in $ml$ across different trials when deployed in real time on the robot, including pours using novel grasps and at novel locations not seen in training.
 
We will show that our estimator allows for very accurate pouring without any sensors external to our robot. To summarize our contributions, this paper is the first to (1) study the use of both joint torque sensors and tactile sensors for pouring tasks; (2) make use of the Jacobian transpose formulation as an approximation of end effector forces during pouring tasks; and (3) demonstrate accurate pouring robust to grasp positions and pouring locations.

\section{Related Work}

At the heart of the robotic pouring task lies the problem of generating a trajectory that regulates the pouring rate to avoid overflow, spillage and, in some cases, achieve a certain target volume at the receiving container. 

A common approach is to learn these trajectories from expert demonstrations\cite{yamaguchi2015pouring, langsfeld2014incorporating}. Rozo et al.\cite{rozo2013force} showed that they can teach a robot to pour drinks from human demonstrations using only force based feedback obtained from a force-torque sensor installed at the end effector of a robot arm. As opposed to our approach, they always pour the same amount (100 ml) and report only binary results as success or failure without evaluating the accuracy of the poured amount. Huang et al.\cite{huang2021robot} uses a motion tracker to collect data from human demonstrations, and a force sensor underneath the receiving container, to develop a model which results in pouring errors between 4 and 12ml. 

Others have tried to leverage fluid simulations to learn pouring trajectories \cite{tamosiunaite2011learning, pan2016robot, do2018learning, kennedy2019autonomous} from synthetic data using deep learning. While these methods circumvent the need to collect data or require human demonstration, they are computationally expensive and may struggle to match the pouring accuracy of closed-loop systems which focus on directly sensing the liquid level or weight. 

Since humans heavily rely on vision to perform pouring tasks, researchers attempted to use different computer vision techniques for robotic pouring \cite{yamaguchi2016stereo, do2016probabilistic} as well as incorporated auxiliary sensing modalities such as audio \cite{liang2019making}. Kennedy et al.\cite{kennedy2019autonomous} demonstrates an extremely accurate system capable of pouring from unknown containers with an error of under 5 ml. However, because their system is meant for wet-labs, they use a highly instrumented and controlled setting for estimating the volume at the receiving container from combined visual feedback and weight measurements. Zhang et al. \cite{zhang2021explainable} uses only visual feedback to decompose the pouring task into multiple hierarchies and build a logical graph which controls the decision process. This work only reports binary pouring results as success or failure without tackling the case of pouring a specific amount. Schenck et al.\cite{schenck2017visual} aims to solve the problem of pouring a specific amount in a general setting using only visual feedback. Similarly to our work, they also use a simple PID controller, but they use a deep network to classify pixels as liquid and then estimate the volume based on these detections. With this approach they demonstrate an average pouring error of 38ml.

In conclusion, to the best of our knowledge, pouring a specific amount of liquid using only proprioceptive sensors or tactile feedback has not been demonstrated so far.
\section{Methodology}

\subsection{Experimental Setup}
\label{sec:exp_setup}

Our experimental setup consists of a Kinova Gen3 robot arm with 7 degrees of freedom with a Robotiq 2F-85 gripper as our manipulator (see Figure \ref{fig:exp_setup}). We replace the original fingers of the gripper with custom designed mounts to hold two BioTac tactile sensors \cite{wettels2008biomimetic}. Because we are interested in the skin deformation at the fingertips, we use the 19 impedance signals delivered by the BioTacs and discard the sensor's AC and DC pressure measurements.

The robot arm is mounted on a tabletop where we perform our experiments. A force-torque sensor (ATI Axia80-M8) is fixed to the table in front of the robot and serves as a base where we place the receiving container. To one side, we also place a 3D printed fixture which allows the robot to find the source container repeatably in the same position. 

A pouring trial begins with the robot in its default resting position. Using pre-programmed waypoints and grasp position, the robot proceeds to grasp the source container, lifts it, and moves it to the pouring position above the receiving container. At this point a pouring controller takes over, and the pouring motion begins using only the rotation of the robot wrist, while the rest of the arm holds position. Once the pouring controller determines that the target weight has been achieved, the wrist rotates back to the initial position with the source container perfectly upright. Finally, the source container is returned to its fixture for the next trial. 

During the pouring trials, our objective is to continuously sense how much weight has left the source container. While collecting data, the force sensor provides this value which we consider our ground truth after adjusting for the time delay that exists between liquid leaving the source container and arriving to the receiving container. The models we train will later replace the force sensor, and the controller will run based on the weight estimates that the model outputs. 

Note that we do not intend to estimate the absolute weight of the held container, but instead we are interested in the weight change experienced at the end effector. We can compute the forces at the end effector in robot base coordinates from joint torques using the Jacobian transpose formulation $\lambda = J^T \tau $ where $\tau$ are the forces applied at the end effector and $\lambda$ represent the joint torque values. Therefore, we can solve for the forces applied at the end effector (``EE Forces'') using the Moore-Penrose pseudo inverse:  $\tau = J^{+T} \lambda$. Note that we obtain a 6 dimensional vector composed of forces and torques across all axes in robot coordinate frame: $\tau = (Fx, Fy, Fz, Tx, Ty, Tz)$

This analytical formulation does not require training data, and we can extract the force component $Fz$ to represent the change in weight at the end effector. We will refer to this method as ``Analytical Fz'' and use it as a baseline to compare our data-driven estimators. Note that we do not want to use the raw joint torques for training, since the learned model would not generalize to new pouring locations. However, because this formulation assumes the manipulator is in static equilibrium, we can only use it as an approximation. 

In the next subsections we describe our pouring controller implementation, the data collection process, and finally the implementation of our poured weight estimators.

\subsection{Pouring Controller}

\begin{figure}[t]
\centering
\includegraphics[clip, trim=5cm 2.1cm 5cm 2cm, width=0.85\linewidth]{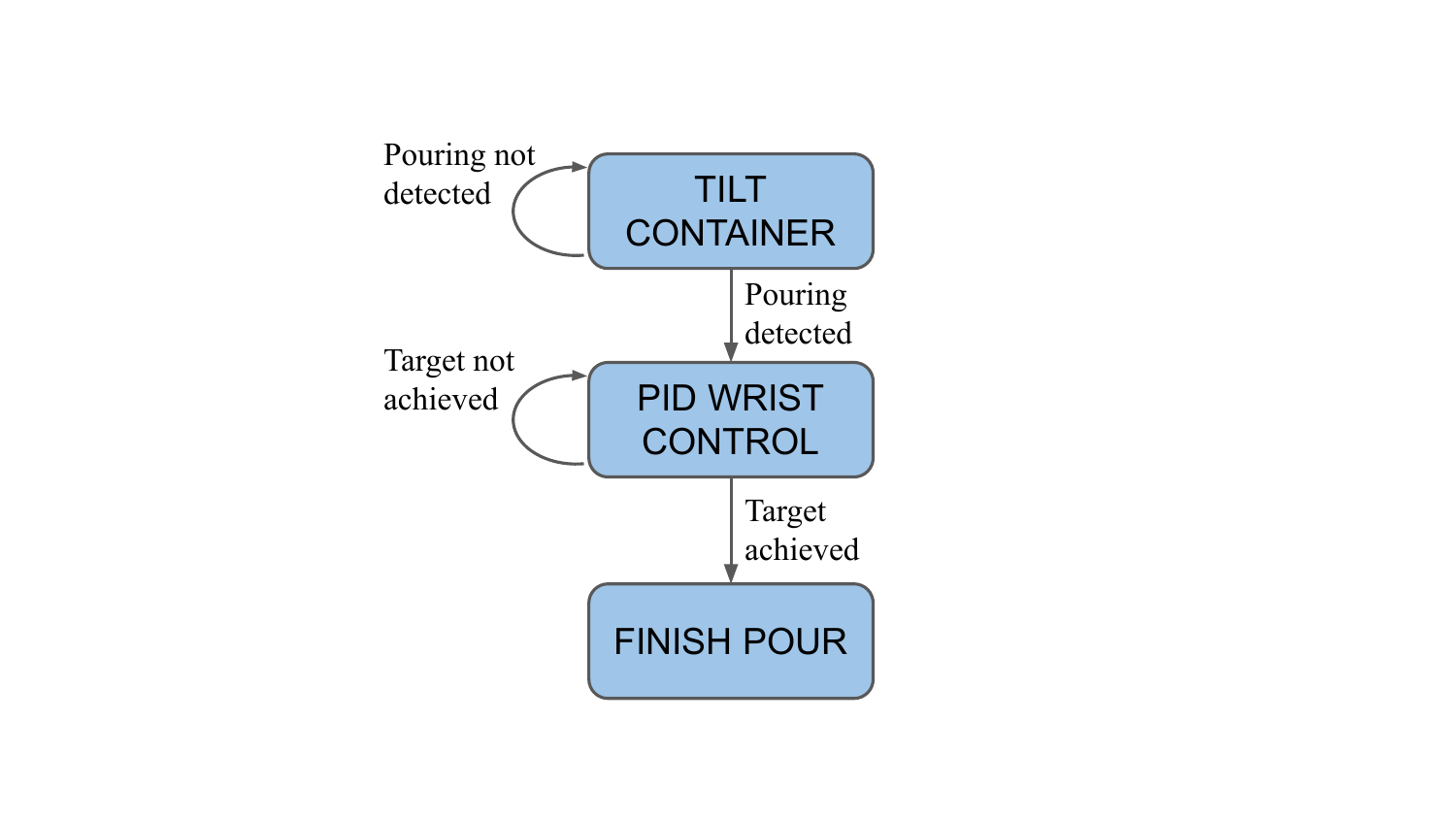}
\caption{Our pouring controller implementation uses 3 distinct states. We initially tilt the container with constant speed until pouring is detected, then enable the PID controller to regulate the wrist velocity until the weight target is achieved}
\label{fig:Controller}
\end{figure}

We design our pouring controller as a finite state machine with three states (Figure \ref{fig:Controller}). The control frequency is determined by the slowest of our sensor feeds, which is the tactile signal from the BioTacs at 100Hz. 

During the first state, a constant velocity is commanded to the robot wrist to begin the pouring. As the wrist rotates, we monitor the signal from the force sensor to determine when the pouring actually begins. To achieve this, we store a window of readings and compute the first derivative of these historic measurements. We declare that the pouring has started once the average value computed over the first derivative exceeds a given threshold. This method provides robustness to the small drift that may occur on the force sensor and the noisy output of our estimators when deployed on the robot. Both the window size and the threshold are tunable parameters. When we detect the pouring has begun, we build a smooth trapezoidal trajectory to be used as reference by the PID controller. This curve is parametrized by a maximum acceleration and velocity, and the initial value corresponds to the latest force measurement. The maximum velocity determines the desired pouring rate in units of N/s. With this trajectory built, we transition to the second state.


During the second state, the PID controller is enabled and becomes the only actor controlling the wrist velocity. At every control iteration, every 10ms, we check whether the target weight has been achieved within a small tolerance value. If so, we transition to the third and final state, where a new trapezoidal trajectory is generated to rotate the wrist back to its upright position and finalize the pour. 

\subsection{Data Collection}

To train our network, we collected 250 pouring trials using the setup described earlier. The same source container and grasp position is used on all these trials. For each trial, we sample each of the following parameters according to a uniform distribution from the interval specified in parenthesis: P gain (0.15 to 0.8), D gain (0 to 0.04), pouring speed (0.1 to 0.8 N/s), weight at source container (200 to 350ml or equivalently 1.96 to 3.92N), and target weight (101 to 306 ml or equivalently 1 to 3N). To ensure pouring trials never results in simply dumping all contents of the source container, the target weight is capped, when necessary, to 75\% of the source container weight.  

Before each trial begins, a trial configuration is sampled, and a human operator fills the source container with the appropriate amount of water using a scale which measures the contents with $\pm$ 1 gram error (we use the known density of water to convert ml to grams). The operator then places the source container in its fixture, and executes the pour sequence, in which the robot arm proceeds to grasp the container, performs the pour, places the container back in its fixture and retracts to start the next trial. 

During each trial, we record training data after the robot has positioned the source container above the receiving container and the actual pouring is about to begin. At this point, we collect a baseline for both the BioTac signals and our computed forces at the end effector. The baseline measurements are simple averages over a 1 second window to further filter out noise, and are subtracted from all subsequent measurements. After recording these baseline measurements we begin logging data for 2 seconds before the controller starts tilting the container and the pouring task begins. The logging of data finishes 4 seconds after the controller deems that we have reached the target weight. 

Data is logged at 100Hz, which is the sampling frequency of the BioTacs, the slowest of all of our sensors. Joint positions, velocities and torques from the robot arm are published at 1 Khz. We configure the force sensor to publish at 200Hz. In the case of the force sensor, we log only the most recent value, which is filtered by the sensor internally with a low pass filter with a cutoff frequency of 1.3Hz. The robot arm data and the tactile data, on the other hand, are stored on a sliding window which represents all measurements over the last 100ms (equivalent to the latest 100 and 10 measurements for the robot arm and BioTacs respectively). To reduce noise in both joint torque sensors and tactile sensors we log the average value computed over this window. For joint positions and velocities we log the most recent values.

\subsection{Poured Weight Estimators}

\begin{figure}[t]
\centering
\includegraphics[clip, trim=6cm 2.5cm 5.5cm 2.5cm, width=0.85\linewidth]{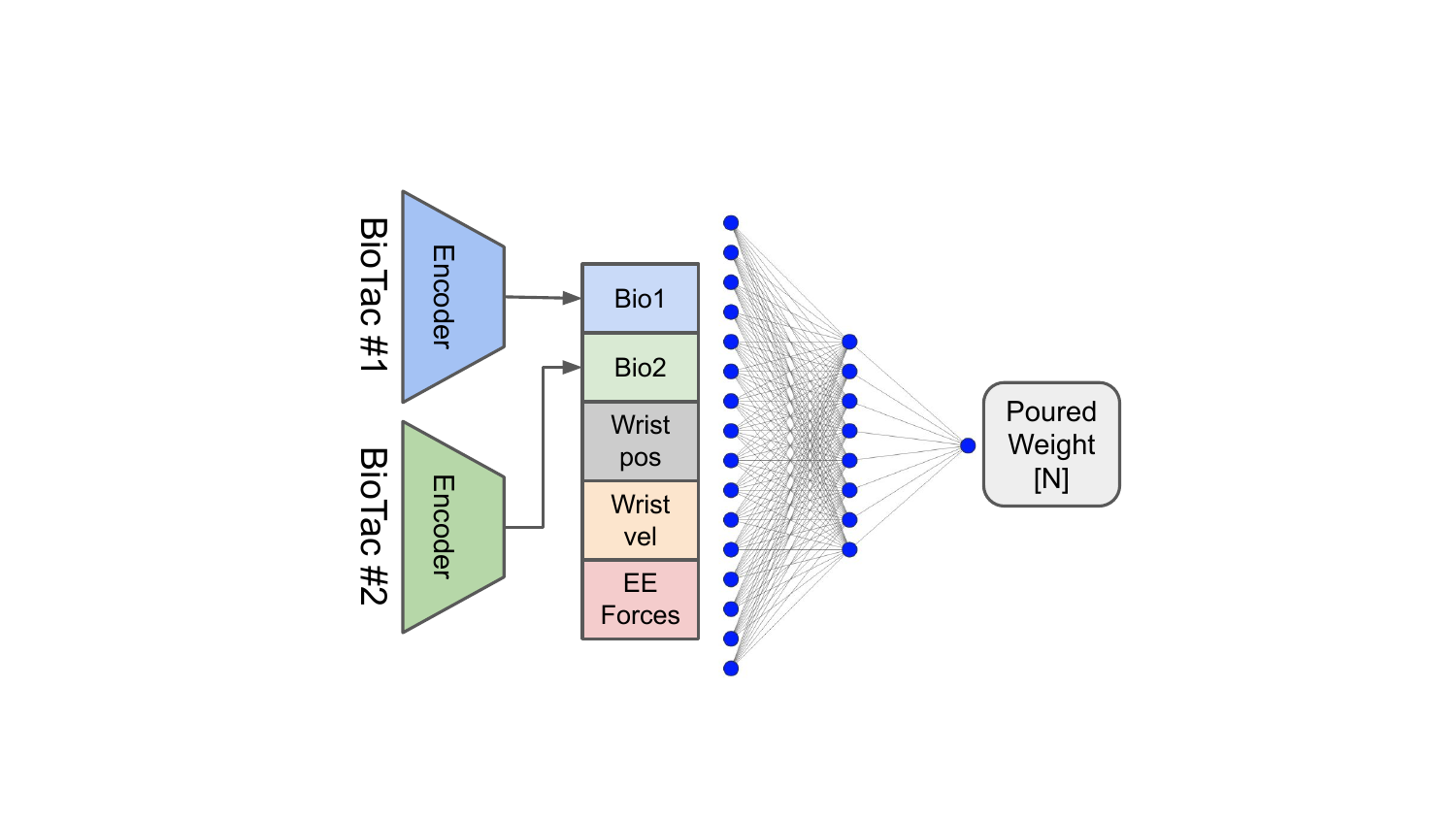}
\caption{We train 3 different estimators to regress the poured weight. Shown here is the \textit{Multimodal} estimator, using both tactile and end-effector (EE) forces as part of the feature input vector. The \textit{Proprioceptive} estimator loses the tactile information, while the \textit{Tactile} estimator loses the EE Forces part of the feature vector}
\label{fig:Estimator}
\end{figure}

Our objective is to use our robot arm proprioception (position, velocity and joint torque sensors) and tactile signals to estimate the weight that has left the source container during a pouring task. Using the data collected as described in the previous subsection, we train several neural networks to regress this quantity.

We will benchmark three different estimators to evaluate the contribution of different sensing modalities. The first estimator will use both tactile and end-effector (EE) forces information. We call it \textit{Multimodal} estimator. The second and third estimators will use tactile and EE forces only respectively and we will refer to them as \textit{Tactile} estimator and \textit{Proprioceptive} estimator. Note that all three estimators make use of wrist position and velocity.

For both the \textit{Multimodal} and \textit{Tactile} estimators, the first stage of our network is a pair of identical encoders to process the impedance signals from each BioTac sensor. These encoders consist of a single linear layer with a ReLU activation function which takes all 19 signals from the BioTac and outputs a vector of 4 dimensions. Note that even though both encoders have the same architecture, they do not share weights and are trained independently. 

The second stage of our network is a simple MLP which processes different feature vectors depending on the estimator. For the \textit{Multimodal} estimator, the encoded tactile signals are concatenated with the wrist position, velocity, and the forces at the end effector computed with the Jacobian transpose method described in section \ref{sec:exp_setup}. The \textit{Tactile} estimator is identical minus the EE forces. Finally the \textit{Proprioceptive} estimator forgoes the tactile information, and the feature vector is composed only of the wrist position and velocity plus the EE forces (see Figure \ref{fig:Estimator})

This results in feature vectors of dimensionality 16, 10 and 8 for the \textit{Multimodal}, \textit{Tactile} and \textit{Proprioceptive} estimators. They become the input to a fully connected network with a single hidden layer of 8, 4 and 4 neurons for each estimator respectively. We use the ReLU activation function except on the output layer which uses a linear transfer function.

The loss function used to train these networks is the mean squared error (MSE) between the predicted value and ground truth. To compute this ground truth we need to adjust the force sensor signal to represent how much weight has left the source container, instead of how much weight has arrived at the receiving container. From the geometry of the source container, the grasp position and the joint angles, we can compute the exact distance from the spout to the receiving container at all given times. We then compute the time it takes for the liquid to travel this distance using a simple free fall model with zero initial velocity. Finally, we shift the recorded data in time to obtain our ground truth signal. This method also circumvents the difficulty of tuning the gains for a system with time-delayed feedback \cite{lee2000pid} when we deploy our estimators to provide feedback in our controller.

To train these estimators we split our data at the trial level, i.e. the data selected for training, validation and testing is comprised of full trials, not by individual datapoints. We randomly chose 210, 20 and 20 trials to be used for training, validation and testing respectively. The training dataset is comprised of 500K datapoints. We train the network for 2000 epochs with a batch size of 5000. We use the Adam optimizer with a fixed step decaying learning rate (initial value of 0.005 which decays with a ratio of 0.7 every 125 epochs).

\section{Evaluation and Results}

\begin{table*}[!th]
\centering
\begin{threeparttable}
\caption{Pouring accuracy trials results - Error [ml]}
\label{tab: accuracy}
\begin{tabular}{lccccccccccccccc}
\hline
\\[-3mm]
\textbf{Estimator}     & \textbf{T1}  & \textbf{T2}  & \textbf{T3}  & \textbf{T4}  & \textbf{T5}  & \textbf{T6}  & \textbf{T7}  & \textbf{T8}  & \textbf{T9}  & \textbf{T10} & \textbf{T11} & \textbf{T12} & \textbf{Avg}    & \textbf{RMSE} & \textbf{Std Dev} \\  \hline 
\\[-3mm] \hline
\\[-2mm]
Analytical Fz & -18 & -14 & -6  & -21 & -19 & -20 & -20 & -10 & -6  & -22 & -24 & -16 & -16.33 &  17.34 & 6.10   \\
Tactile       & 2   & 94  & 60  & 40$^*$ & 40$^*$ & 40$^*$ & 76  & 70  & 104 & 60$^*$ & 60$^*$ & 60$^*$ & 58.83  &  64.25 & 26.96  \\
Proprioceptive      & -13 & -11 & -10 & -11 & -13 & -13 & -15 & -11 & -12 & -13 & -13 & -11 & -12.17 &  12.24 & 1.40   \\
Multimodal    & 3   & 8   & 8   & -5  & 6   & 14  & 8   & 1   & 8   & -3  & 10  & 10  & 5.67   &  7.81  & 5.61  

\end{tabular}
\begin{tablenotes}
\item[*] all contents were poured out of the source container
\end{tablenotes}
\end{threeparttable}
\end{table*}

Our objective is to replace the force sensor used during training with a regressor which outputs the current weight poured out of the source container. In the previous section we described three different estimators to be trained using either only tactile, only end effector forces or both modalities. We also propose the use of an analytical formulation where we take the $z$ component of the computed forces at the end effector in robot coordinates as a baseline comparison. 

Our first evaluation of these methods consists in comparing their individual predictions against ground truth on our collected test set, composed of 20 randomly selected trials. The main metric for comparison is the mean square error (MSE) between the estimator output and the ground truth signal aggregated over all test trials.

\begin{figure}[b]
\centering
\includegraphics[clip, trim=2cm 0.25cm 2cm 1.5cm, width=0.98\linewidth]{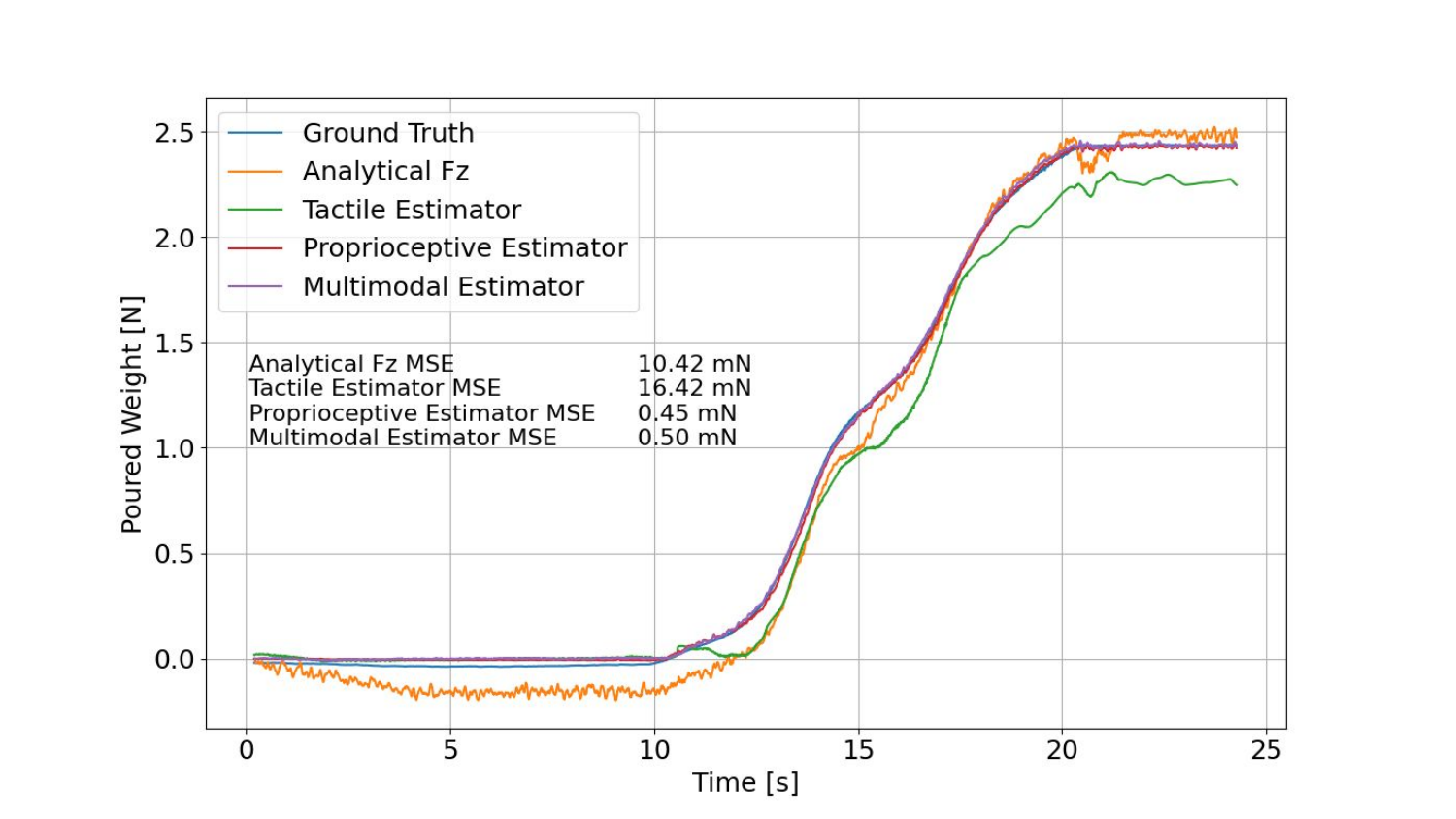}
\caption{Pouring curves of a single pouring trial from our test set comparing the prediction of our estimators against ground truth and the analytical $Fz$ component which we extract from our computation of EE Forces}
\label{fig:estimators_mse}
\end{figure}

The baseline method of computing the change in $Fz$ through the Jacobian transpose yields a MSE of 8.06 mN. The MSE for the \textit{Tactile}, \textit{Proprioceptive} and \textit{Multimodal} estimator were 7.81 mN, 0.62 mN and 0.52 mN respectively. Figure \ref{fig:estimators_mse} shows one of the test trials with traces for each one of our estimators and baseline against ground truth with MSE values for that particular trial alone. The \textit{Analytical Fz} method presents some noise during the initial container tilt stage before the pouring starts but then follows the ground truth curve reasonably well. The \textit{Tactile} estimator correctly predicts the beginning of the pour but shows larger error during the actual pouring stage and on the final predicted value. The \textit{Proprioceptive} estimator makes good use of the extra information of all forces felt at the end-effector (including torques) plus wrist position and velocity to improve on the performance of the \textit{Analytical Fz}. The \textit{Multimodal} estimator shows a marginal improvement with the addition of tactile information. These trends are visible across most test trials. 

Our second evaluation consists of deploying each of our estimators, plus our analytical baseline, on our robot arm and performing a set of representative pouring trials, both within and outside the training data distribution. 

First we tackle the performance of our models within the distribution that they have been trained on. This means that the testing parameters are within those used to collect our training data. Table \ref{tab: eval_trials} describes 12 instances of pouring trials with various parameters that sweep through different pouring speeds, liquid volume at the source container and target volume to achieve on the receiving container. For each one of our estimators and baseline, we perform these pouring trials on our robot and record the pouring accuracy as the error between the actual quantity poured during the trial versus the target as defined by the trial parameters on Table \ref{tab: eval_trials}. After each trial we weight the receiving container and using the known density of water we report the error in milliliters (a positive error value means we poured more liquid than our target). Results are shown in Table \ref{tab: accuracy}.

\begin{table}[t]
\centering
\caption{Evaluation trials parameters}
\label{tab: eval_trials}
\begin{tabular}{lccc}
\hline
\\[-3mm]
\textbf{Trial \#} & \begin{tabular}[c]{@{}c@{}}\textbf{Pour speed} \\ {[}\textbf{ml/s}{]}\end{tabular} & \begin{tabular}[c]{@{}c@{}}\textbf{Source volume}\\ {[}\textbf{ml}{]}\end{tabular} & \begin{tabular}[c]{@{}c@{}}\textbf{Target volume}\\ {[}\textbf{ml}{]}\end{tabular} \\ \hline
\\[-3mm] \hline
\\[-2mm]
1  & 15  & 220  & 120 \\
2  & 45 & 220 & 120 \\
3  & 80 & 220 & 120 \\
4  & 15 & 220 & 180 \\
5  & 45 & 220 & 180 \\
6  & 80 & 220 & 180 \\
7  & 15 & 320 & 150 \\
8  & 45 & 320 & 150 \\
9  & 80 & 320 & 150 \\
10 & 15 & 320 & 260 \\
11 & 45 & 320 & 260 \\
12 & 80 & 320 & 260                                    
\end{tabular}
\end{table}

\begin{figure}[t]
\centering
\includegraphics[clip, trim=3cm 0.5cm 3cm 1.6cm, width=0.80\linewidth]{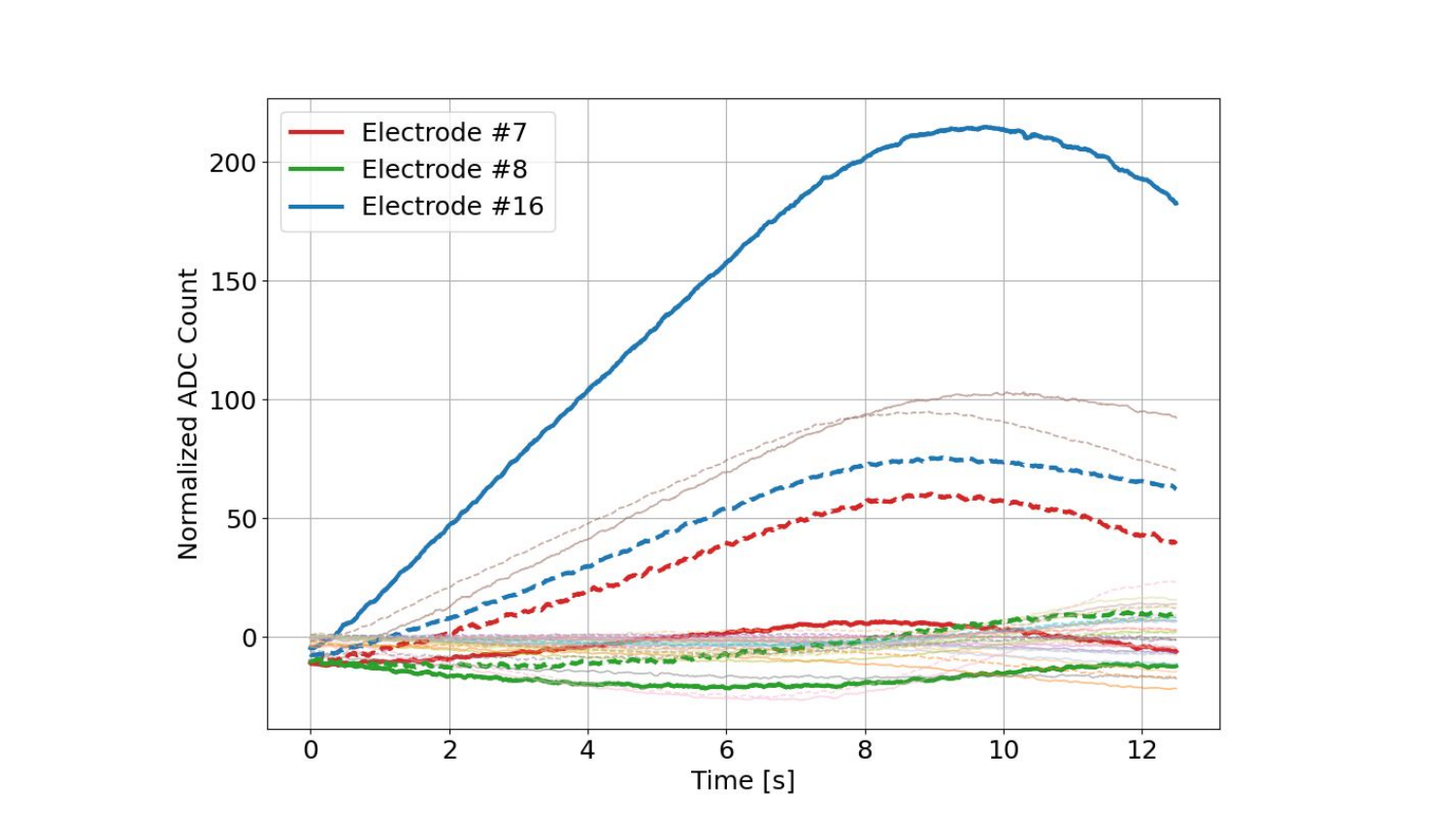}
\caption{Tactile signal variation for a 1.5 mm off-center grasp disturbance. We highlight 3 electrodes which show the largest discrepancies}
\label{fig:tact_variability}
\end{figure}

The first thing to note is that our \textit{Tactile} estimator performance diverges notably from the results obtained with our off-line evaluation, to the point of failure to perform the task. We investigated the cause of this divergence and found that our tactile signals present very high sensitivity to small grasp perturbations (1-2mm off-center), yielding fairly different patterns and signal amplitudes as a result. In our experimental setup, the source container fixture tolerance is 1mm which stacks on top of our robot arm repeatability, which the manufacturer does not guarantee for the quick-connect version of the arm \cite{kinova}, but can be considered to be 1mm at best (metric they provide for the fixed-base version of the arm). Figure \ref{fig:tact_variability} depicts the difference in tactile signals for an open loop pour (same trajectory) when we perturb the grasp to be off-center by 1.5mm.

While casual inspection might suggest that the \textit{Multimodal} estimator outperforms the \textit{Proprioceptive} because of its lower root mean squared error (RMSE), there is a clear bias on the latter towards under filling the receiving container. This manifests as a systematic error of around -11 ml. If we ``calibrate'' for this offset, this estimator is incredibly consistent. The \textit{Analytical Fz} estimator presents a similar behaviour, but with a larger error spread, resulting in a standard deviation five times larger than the \textit{Proprioceptive} estimator. The addition of tactile is likely reducing the performance of the \textit{Multimodal} estimator compared to the \textit{Proprioceptive} estimator, increasing the spread of the prediction error, due to the issues previously discussed with our tactile signals.

\begin{figure}[t]
\centering
\includegraphics[clip, trim=2.2cm 0cm 2.2cm 0.25cm, width=0.90\linewidth]{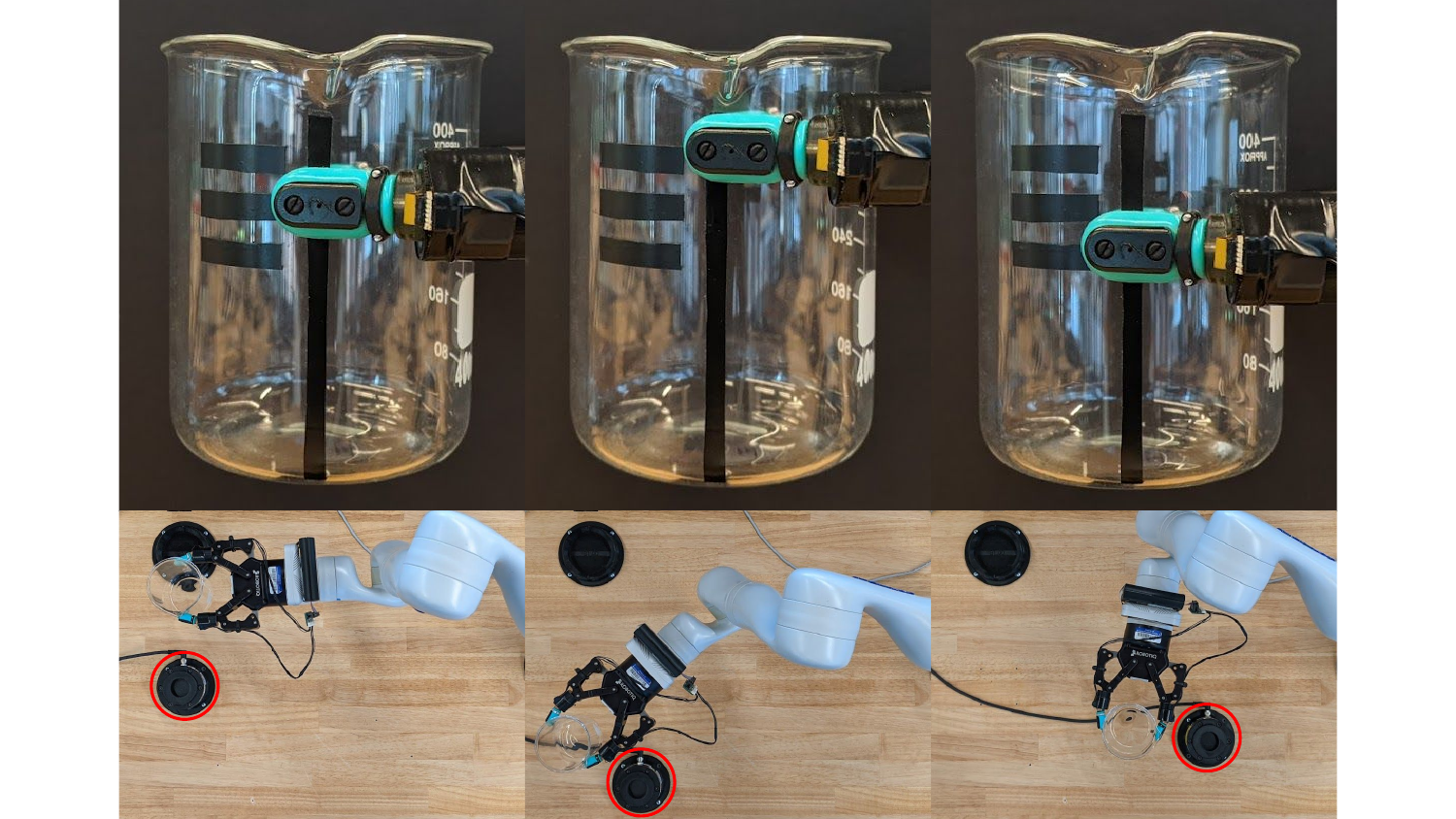}
\caption{Left pictures show the grasp and location we trained our estimators on. We evaluate their generalization performance by testing them on novel grasp positions and novel pouring locations (middle and right).}
\label{fig:generalization}
\end{figure}

We would also like to evaluate how robust these estimators might be when operating outside the training distribution. We are particularly interested in two cases: pouring at a different location, and pouring using a different grasp on the source container. Two novel locations and two novel grasps are introduced as shown in Figure \ref{fig:generalization}. For this test we select trials 1,6,8 and 12 and we perform these at each of the 2 novel locations and then using the 2 novel grasps (at the trained-on location). The pouring accuracy results are shown in Table \ref{tab: novel_trials}. Using novel grasps results in a small performance drop across the board, although similar trends to those observed before remain: the \textit{Analytical Fz} and \textit{Proprioceptive} estimator still present a constant bias that results in underfilling the container (individual trial results not shown). For the case of novel locations, both estimators show a larger error spread to go along with their inherent bias, which is to be expected as these are locations never seen in training. Once again, the \textit{Tactile} estimator fails to perform the task and the addition of tactile signals does not improve performance for the \textit{Multimodal} estimator.

\begin{table}[t]
\centering
\caption{Novel location and grasp trial results [ml]}
\label{tab: novel_trials}
\begin{tabular}{lcccc}
\hline
\\[-3mm]
\multirow{2}{*}{\textbf{Estimator}} & \multicolumn{2}{c}{\textbf{Novel locations}} & \multicolumn{2}{c}{\textbf{Novel grasps}} \\
                                    & \textbf{RMSE}       & \textbf{Std Dev}       & \textbf{RMSE}      & \textbf{Std Dev}     \\ \hline
\\[-3mm] \hline
\\[-2mm]
Analytical Fz                       & 10.71               & 9.20                   & 21.53              & 8.84                 \\
Tactile                             & 61.50               & 20.69                  & 72.17              & 40.16                \\
Proprioceptive                      & 9.70                & 7.04                   & 13.81              & 4.58                 \\
Multimodal                          & 11.42               & 5.13                   & 8.51               & 9.09                
\end{tabular}
\end{table}

\section{Conclusions}

In this paper we focused on the problem of pouring a specific amount of liquid into a receiving container using only the robot's own tactile and proprioceptive sensors, without any environmental instrumentation. We avoid the use of cameras because we are mostly interested in a pouring solution that can be deployed in a general scenario, without strict controls on lighting or background conditions.

During this study we have demonstrated that we can leverage a robot arm's joint torque sensors to effectively sense the weight change of a held container during a pouring task. Our trained \textit{Proprioceptive} model can pour within $\approx 10 ml$ on average from the target, a result equivalent to other state of the art methods which rely on external sensors. Our approach shown to be robust enough to pour at novel locations and with different grasps on the source container. 

On the flip side, we have debunked our initial hypothesis regarding the use of tactile sensors for pouring tasks. Our results show that the variability associated with the signals of our BioTac sensors with respect to minor grasp perturbances make it very challenging to train a regressor that provides stable and reliable predictions over time. While we believed that tactile sensors would bridge the gap in performance between computing forces at the end effector through joint torques versus a dedicated force-torque sensor, it turns out we underestimated the accuracy of joint torque sensors and overestimated the usefulness of the tactile modality. 

Our approach is not without its limitations. We assume prior knowledge of the state of the receiving container in terms of its capacity and current liquid level, since we do not directly sense these parameters to avoid spillage. In our trials, the receiving container is always assumed to be both empty and large enough to avoid overfilling. In similar fashion, we also do not sense or model the amount of liquid in the source container, which results in our trials starting with an open-loop tilt motion until we detect pouring has started. To avoid an initial large pouring rate, we require this initial rotation of the source container to happen at a low speed, which in turn increases the total time needed to complete the full pouring task. Our training data is only collected with water, hence it is not clear how well this method might generalize to granular media or high viscosity liquids. Finally, we did not investigate the use of different source containers and their effect on pouring performance. 

We believe the results presented in this paper are relevant because, to our knowledge, there have not been any previous studies that use joint torque measurements or tactile signals at the fingertips to demonstrate pouring skills. While a similar framework could be proposed with a force-torque sensor at the wrist, as it has been done in the literature, we believe it is valuable to study the feasibility of replacing these specialized sensors when joint torques are available to facilitate integration, free up available payload on the robot arm, eliminate external wiring, and reduce cost. While the tactile modality did not provide significant additional utility in our experiments, it is possible that other tactile sensors or processing methods could extract further performance to enable even more precise and robust pouring. 


\bibliographystyle{IEEEtran}
\bibliography{biblio}

\end{document}